\documentclass[conference]{IEEEtran}
\IEEEoverridecommandlockouts
\usepackage{cite}
\usepackage{amsmath,amssymb,amsfonts}
\usepackage{algorithmic}
\usepackage{graphicx}
\usepackage{textcomp}
\usepackage{xcolor}
\usepackage{booktabs}
\usepackage{makecell}
\usepackage{multirow}
\usepackage{arydshln}
\usepackage{lineno,hyperref}
\def\BibTeX{{\rm B\kern-.05em{\sc i\kern-.025em b}\kern-.08em
    T\kern-.1667em\lower.7ex\hbox{E}\kern-.125emX}}
\begin{document}

\title{DDFAV: Remote Sensing Large Vision Language Models Dataset and Evaluation Benchmark\\
}

\author{\IEEEauthorblockN{Haodong Li}
\IEEEauthorblockA{\textit{Liaoning Technical University} \\
\textit{School of Software}\\
Huludao, China \\
472321734@stu.lntu.edu.cn}
\and
\IEEEauthorblockN{Haicheng Qu*}
\IEEEauthorblockA{\textit{Liaoning Technical University} \\
\textit{School of Software}\\
Huludao, China \\
quhaicheng@lntu.edu.cn}
\and
\IEEEauthorblockN{Xiaofeng Zhang}
\IEEEauthorblockA{\textit{Shang Hai Jiao Tong University} \\
\textit{department of automation}\\
Shanghai, China \\
framebreak@sjtu.edu.cn}
}

\maketitle

\begin{abstract}
With the rapid development of large vision language models (LVLMs), these models have shown excellent results in various multimodal tasks. Since LVLMs are prone to hallucinations and there are currently few datasets and evaluation methods specifically designed for remote sensing, their performance is typically poor when applied to remote sensing tasks. To address these issues, this paper introduces a high-quality remote sensing LVLMs dataset, DDFAV, created using data augmentation and data mixing strategies. Next, a training instruction set is produced based on some high-quality remote sensing images selected from the proposed dataset. Finally, we develop a remote sensing LVLMs hallucination evaluation method RSPOPE based on the proposed dataset and evaluate the zero-shot capabilities of different LVLMs. Our proposed dataset, instruction set, and evaluation method files are available at \href{https://github.com/HaodongLi2024/rspope}{https://github.com/HaodongLi2024/rspope}.

\end{abstract}


\section{Introduction}

\begin{figure}[hbtp]
    \vspace{-0.5cm}
    \centering
    \includegraphics[width=8.8cm]{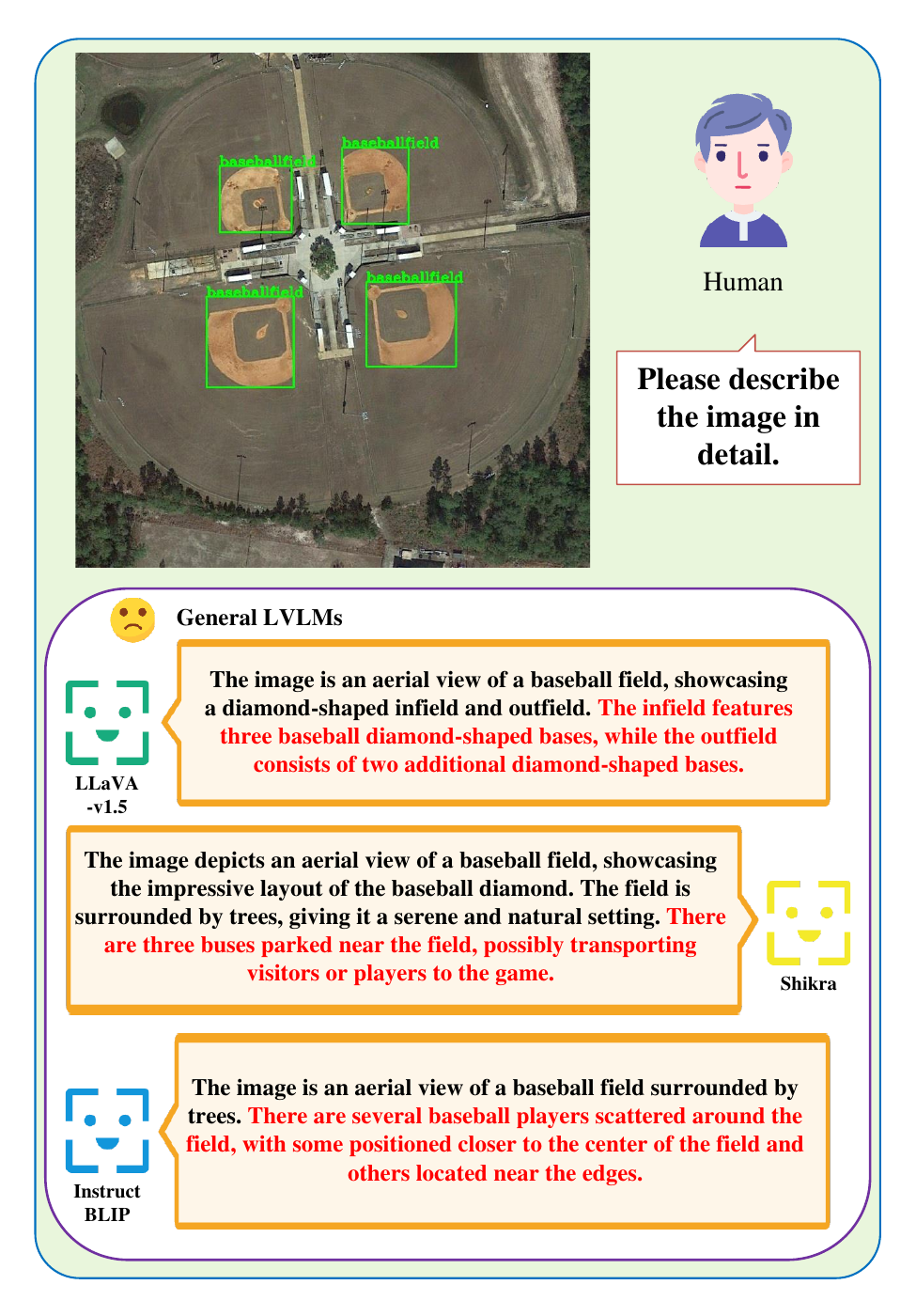} 
    \caption{Comparison of hallucination phenomena of general LVLMs in image captioning tasks for remote sensing images.}
    \label{fig:general_remote}
    \vspace{-0.5cm}
\end{figure}

Remote sensing technology uses image data acquired from high altitudes to analyze and understand various phenomena on the Earth's surface. It is widely applied in fields such as land cover classification, disaster monitoring, and environmental protection. Recently, with the successful application of large vision language models (LVLMs) in natural scenes, researchers have begun to introduce these models into the field of remote sensing to achieve breakthroughs in tasks like image caption generation~\cite{zhao2021high}, scene classification~\cite{liu2024remoteclip}, and complex reasoning~\cite{wang2024earthvqa}. However, general LVLMs designed for natural scenes tend to produce many hallucinations~\cite{ghosh2024vdgd,an2024agla,huang2024opera,qu2024alleviating,liu2024paying,wu2024evaluating,kim2024code,xiao2024detecting,nips}, and there are significant differences between remote sensing images and natural images. These differences are evident not only in the complexity of image content and target size but also in the requirement for complex spatial reasoning from a unique bird's-eye perspective in remote sensing images. These factors often lead LVLMs to produce hallucinations in remote sensing tasks, as shown in Fig~\ref{fig:general_remote}, revealing the limitations of existing remote sensing vision-language models and datasets.

Most existing remote sensing LVLMs datasets are either restricted to a single task ~\cite{zhang2023rs5m,yang2010bag} or lack diversity and detail~\cite{lu2017exploring}, limiting the models' generalization and multi-task processing capabilities. In addition, the annotation quality of many datasets is inconsistent, there is a lack of complex scene reasoning data~\cite{qu2016deep}, and the image caption instruction set caption is too short~\cite{zhang2014saliency}, which leads to significant bias or errors in the training models when processing diverse remote sensing images. Therefore, there is an urgent need for a high-quality remote sensing visual language dataset that covers a wider range of scenes, perspectives, and categories and can perform multiple tasks, ranging from simple image description to complex reasoning.

\begin{figure*}[hbtp]
    \centering
    \includegraphics[height=7.1cm,width=0.8\textwidth]{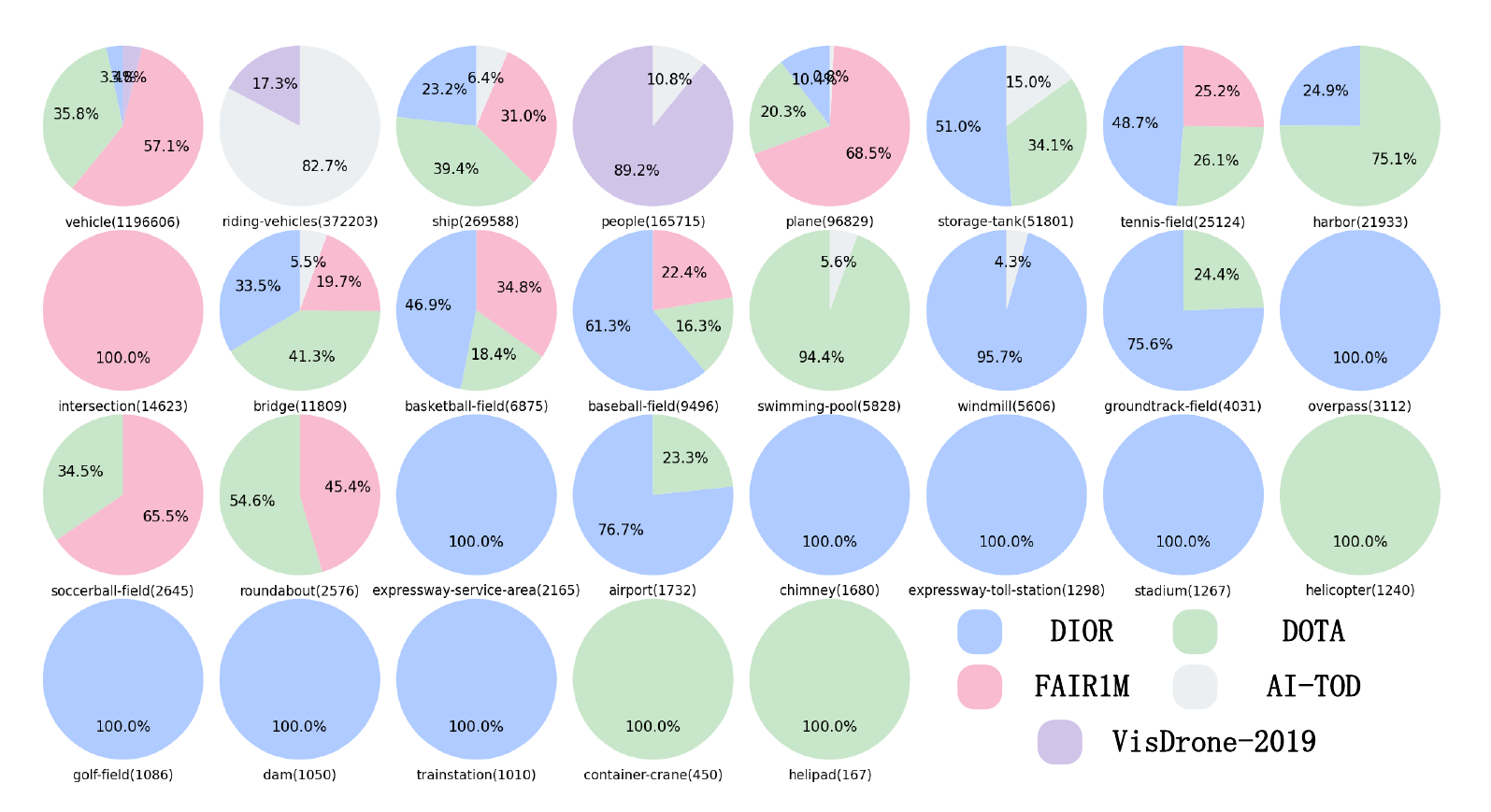} 
    \caption{The number of object categories and source distribution of our proposed remote sensing LVLMs dataset DDFAV.}
    \label{fig:geo_lvlm}
    \vspace{-0.5cm}
\end{figure*}

Current evaluation methods for remote sensing LVLMs also have notable shortcomings. For example, in image captioning tasks, most evaluation methods focus on the overall similarity between the generated text and the reference text, while overlooking the ability of LVLMs to accurately recognize specific objects in remote sensing images, such as small or overlapping objects. Metrics like BLEU~\cite{papineni2002bleu} and METEOR~\cite{banerjee2005meteor} emphasize n-gram overlap between generated and reference captions, ROUGE-L~\cite{lin2004rouge} focuses on the longest common subsequence between texts, and CIDEr-D~\cite{vedantam2015cider} measures consistency across multiple reference captions. For visual question answering tasks, methods such as RSVQA~\cite{lobry2020rsvqa} include four types of questions: existence, comparison, rural/urban classification, and counting, seemingly providing a comprehensive evaluation of remote sensing LVLMs. However, the subjectivity and variability introduced by open-ended questions present a significant challenge for achieving consistent and reliable evaluation results. The main contributions of this paper can be
summarized as follows:
\begin{itemize}
    \item We analyze the shortcomings of existing remote sensing LVLMs on dedicated datasets and propose a multi-category, multi-view, and more uniformly scaled remote sensing LVLMs dataset, DDFAV, created using data enhancement and data mixing techniques.
    \item Using the proposed DDFAV dataset, we select a subset to create an instruction set for training remote sensing LVLMs, covering tasks such as image captioning, visual question answering, and complex reasoning.
    \item We develop an RSPOPE evaluation method, based on POPE~\cite{li2023evaluating}, for current remote sensing LVLMs using the proposed DDFAV dataset and employ mainstream LVLMs to evaluate their zero-shot remote sensing image recognition capabilities.

\end{itemize}

\section{Method}

\subsection{DDFAV Remote Sensing LVLMs Dataset}

\begin{figure*}[hbtp]
    \centering
    \includegraphics[width=0.8\textwidth]{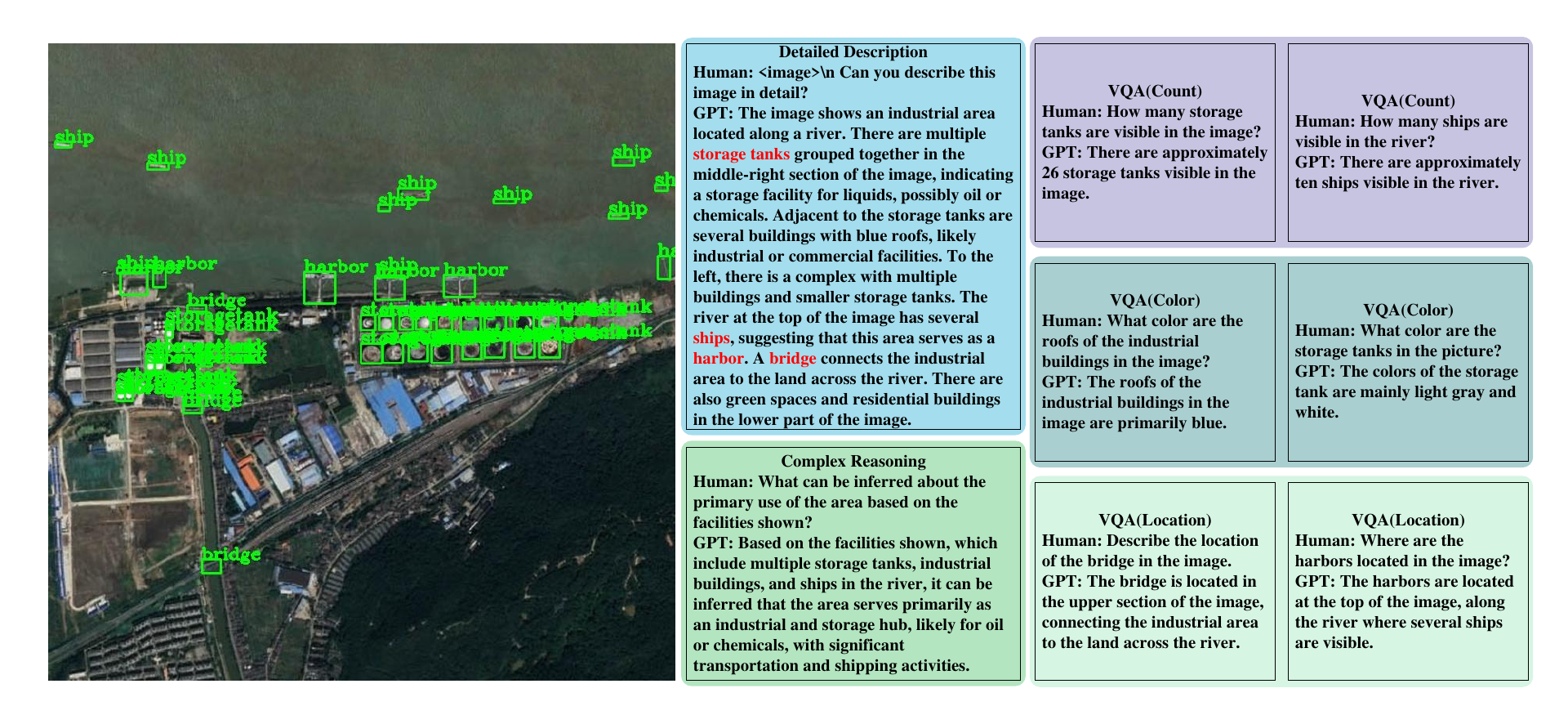} 
    \caption{An example from the DDFAV remote sensing LVLMs instruction set includes a remote sensing image with 8 question-answer pairs: 1 detailed image description, 1 complex reasoning question, 2 visual questions about color, 2 visual questions about counting, and 2 visual questions about object location.}
    \label{fig:instruction_set}
    \vspace{-0.5cm}
\end{figure*}

The remote sensing LVLMs dataset DDFAV we propose includes five target detection datasets: DIOR~\cite{li2020object}, DOTA~\cite{xia2018dota}, FAIR1M~\cite{sun2022fair1m}, VisDrone-2019~\cite{du2019visdrone}, and AI-TOD~\cite{wang2021tiny}. The first three are SAMRS~\cite{wang2024samrs} remote sensing image target detection datasets used by GeoChat. These three datasets contain a wealth of object categories (such as cars, ships, airplanes, various venues, etc.) and standard perspectives from remote sensing satellite imagery. However, these three datasets have two shortcomings. The first is the lack of remote sensing object information, such as pedestrians and riding vehicles. To address this, we add a fourth dataset, the VisDrone-2019 dataset, which was publicly released by the AISKYEYE team at Tianjin University. This dataset not only compensates for the missing information on objects like pedestrians and riding vehicles but also provides remote sensing images from the perspective of drone aerial photography, making the dataset more diverse. Second, the sizes of objects in the first three datasets are relatively uniform. A major challenge in remote sensing image target detection is the low accuracy of small object detection. This issue is also evident when observing the hallucination phenomenon in LVLMs. To address this, we add a fifth dataset, the AI-TOD dataset proposed by Wuhan University in 2020, where 87.7\% of the objects are smaller than 32×32 pixels. The mean and standard deviation of the object sizes are 12.8 pixels and 5.9 pixels, respectively, which are much smaller than those in other remote sensing and aerial image datasets. Finally, our proposed DDFAV dataset contains 29 remote sensing object categories, including large objects such as airports, stadiums, bridges, and small objects such as cars, pedestrians, and ships. It also covers different scenes in cities and rural areas as well as different perspectives of satellites and drones. The object categories and number information of the dataset are shown in Fig.~\ref{fig:geo_lvlm}.

\subsection{DDFAV Remote Sensing LVLMs Instruction Set}
Additionally, we create corresponding instruction sets for high-quality remote sensing images selected from the proposed DDFAV dataset by combining model-generated annotations with manual quality checks. The statistics of our proposed instruction set are shown in Table~\ref{tab:dataset_statistics}. First, we input each image into GPT-4o, which generates 8 question-answer pairs related to the image. The answers provide detailed descriptions, complex reasoning, and additional information such as the number of objects, their colors, sizes, positions in the image, and relationships between objects. Then, through manual inspection and modification, we align these details with the true content of the image to obtain the final instruction set. An example of the instruction set we propose is shown in Fig~\ref{fig:instruction_set}.
\vspace{-0.5cm}

\begin{table}[h]
\caption{DDFAV instruction set information statistics.}
\label{tab:dataset_statistics}
\centering
\renewcommand{\arraystretch}{1} 
\begin{tabular}{l|c|c|c|c}
\hline
\textbf{Dataset} & \textbf{Images} & \textbf{Sentences} & \textbf{Words} & \textbf{Caption length} \\
\hline
DIOR & 499 & 7636 & 127951 & 46.81 \\
DOTA & 673 & 10756 & 226218 & 85.22 \\
FAIR1M & 167 & 2704 & 49676 & 74.22 \\
AI-TOD & 37 & 564 & 10961 & 71.73 \\
VisDrone-2019 & 379 & 5990 & 92105 & 44.13 \\
\hline
\end{tabular}
\vspace{-0.3cm}
\end{table}

\begin{figure}[t]
    \vspace{-0.3cm}
    \centering
    \includegraphics[width=8.8cm]{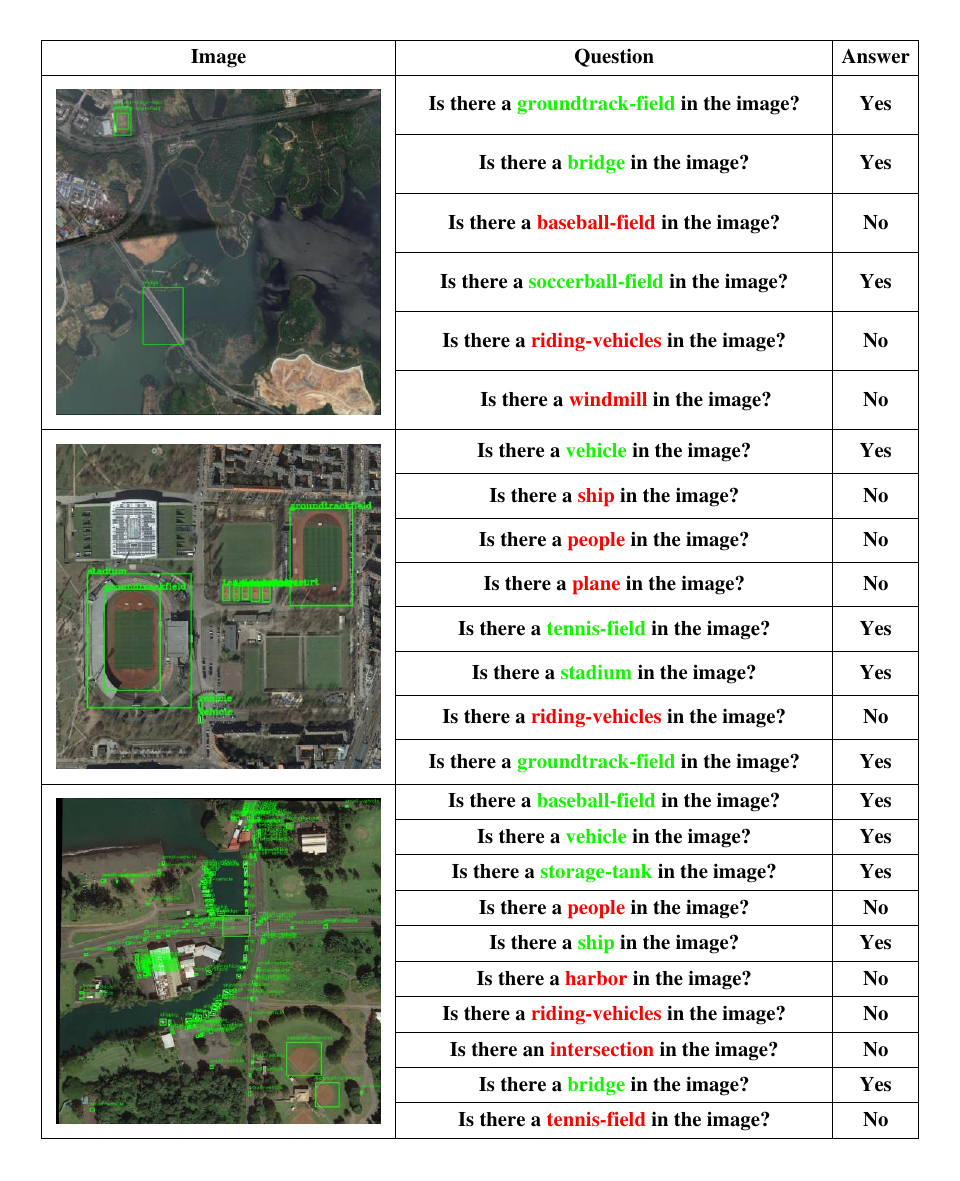} 
    \caption{The RSPOPE evaluation benchmark is designed for specific images in the DDFAV dataset. From top to bottom, the settings of the three pictures are easy + random, medium + popular, and hard + adversarial.}
    \label{fig:RSPOPE}
    \vspace{-0.3cm}
\end{figure}

\begin{table*}[htbp]
\caption{The proposed RSPOPE evaluation method conducts zero-shot evaluations on 7B-sized LVLMs using the selected DDFAV dataset.}
\label{general_table}
\centering
\begin{tabular}{lllcccc|c|cc}
\hline
\textbf{Level} & \textbf{Setting} & \textbf{Model} & \textbf{LLM} & \textbf{Accuracy} & \textbf{Precision} & \textbf{Recall} & \textbf{F1 Score} & \textbf{Yes (\%)} & \textbf{Time}\\
\hline
\multirow{18}{*}{\textbf{Easy}} & \multirow{6}{*}{\textit{Random}} & Minigpt4~\cite{zhu2023minigpt} & Vicuna-v0-7B~\cite{chiang2023vicuna} & 68.17 & 67.14 & 98.44 & 79.83 & 93.83 & 06:30 \\
 & & Minigpt4 & Llama2-7B~\cite{touvron2023llama} & 65.50 & 66.30 & 93.75 & 77.67 & 90.50 & 06:48 \\
 & & InstructBLIP~\cite{NEURIPS2023_9a6a435e} & Vicuna-v1.1-7B & 74.67 & \textbf{84.73} & 73.70 & 78.83 & 55.67 & 06:48\\
 & & Shikra~\cite{chen2023shikra} & Vicuna-7B & 71.00 & 72.44 & 88.28 & 79.58 & 78.00 & 21:34 \\
 & & LLaVA-v1.5~\cite{liu2024visual} & Vicuna-v1.5-7B & 71.00 & 68.95 & \textbf{99.48} & 81.45 & 92.33 & 13:19 \\
 & & GeoChat\cite{kuckreja2024geochat} & Vicuna-v1.5-7B & \textbf{75.00} & 74.89 & 91.67 & \textbf{82.44} & 78.33 & 18:16 \\
\cline{2-10}
 & \multirow{6}{*}{\textit{Popular}} & Minigpt4 & Vicuna-v0-7B & 68.17 & 67.38 & 97.40 & 79.66 & 92.50 & 06:36\\
 & & Minigpt4 & Llama2-7B & 71.50 & 71.01 & 93.75 & 80.81 & 84.50 & 06:48 \\
 & & InstructBLIP & Vicuna-v1.1-7B & 75.00 & 85.24 & 73.70 & 79.05 & 55.33 & 06:53 \\
 & & Shikra & Vicuna-7B & 82.17 & 84.54 & 88.28 & 86.37 & 66.83 & 21:26 \\
 & & LLaVA-v1.5 & Vicuna-v1.5-7B & 73.83 & 71.14 & \textbf{99.48} & 82.95 & 89.50 & 13:52 \\
 & & GeoChat & Vicuna-v1.5-7B & \textbf{89.67} & \textbf{92.15} & 96.17 & \textbf{91.91} & 63.67 & 18:30 \\
\cline{2-10}
 & \multirow{6}{*}{\textit{Adversarial}} & Minigpt4 & Vicuna-v0-7B & 66.50 & 66.19 & 97.40 & 78.82 & 94.17 & 06:33 \\
 & & Minigpt4 & Llama2-7B & 67.50 & 67.80 & 93.75 & 78.69 & 88.50 & 06:47\\
 & & InstructBLIP & Vicuna-v1.1-7B & 67.50 & 75.07 & 73.70 & 74.38 & 62.83 & 06:44 \\
 & & Shikra & Vicuna-7B & 75.83 & 77.22 & 88.28 & 82.38 & 73.17 & 21:26 \\
 & & LLaVA-v1.5 & Vicuna-v1.5-7B & 69.50 & 67.85 & \textbf{99.48} & 80.68 & 93.83 & 13:50 \\
 & & GeoChat & Vicuna-v1.5-7B & \textbf{78.83} & \textbf{78.75} & 91.67 & \textbf{84.72} & 74.50 & 17:54 \\
 \hline
\multirow{18}{*}{\textbf{Medium}} & \multirow{6}{*}{\textit{Random}} & Minigpt4 & Vicuna-v0-7B & 49.50 & 49.27 & 94.15 & 64.69 & 93.88 & 08:40 \\
 & & Minigpt4 & Llama2-7B & 49.50 & 49.24 & 90.33 & 63.73 & 90.13 & 09:03 \\
 & & InstructBLIP & Vicuna-v1.1-7B & \textbf{73.00} & \textbf{72.41} & 72.77 & 72.59 & 49.38 & 08:53 \\
 & & Shikra & Vicuna-7B & 66.13 & 60.97 & 86.26 & 71.44 & 69.50 & 28:41 \\
 & & LLaVA-v1.5 & Vicuna-v1.5-7B & 59.88 & 55.07 & \textbf{99.49} & 70.90 & 88.75 & 18:23 \\
 & & GeoChat & Vicuna-v1.5-7B & 69.63 & 63.84 & 88.04 & \textbf{74.01} & 67.75 & 24:21 \\
\cline{2-10}
 & \multirow{6}{*}{\textit{Popular}} & Minigpt4 & Vicuna-v0-7B & 54.88 & 52.26 & 94.15 & 67.21 & 88.50 & 08:39\\
 & & Minigpt4 & Llama2-7B & 59.25 & 55.21 & 90.33 & 68.53 & 80.38 & 09:05 \\
 & & InstructBLIP & Vicuna-v1.1-7B & 81.38 & 87.20 & 72.77 & 79.33 & 41.00 & 09:13 \\
 & & Shikra & Vicuna-7B & 81.13 & 77.75 & 86.26 & 81.79 & 54.50 & 28:39 \\
 & & LLaVA-v1.5 & Vicuna-v1.5-7B & 67.38 & 60.15 & \textbf{99.49} & 74.98 & 81.25 & 18:19 \\
 & & GeoChat & Vicuna-v1.5-7B & \textbf{89.88} & \textbf{91.05} & 88.04 & \textbf{89.52} & 47.50 & 24:30 \\
\cline{2-10}
 & \multirow{6}{*}{\textit{Adversarial}} & Minigpt4 & Vicuna-v0-7B & 52.13 & 50.68 & 94.15 & 65.89 & 91.25 & 08:40 \\
 & & Minigpt4 & Llama2-7B & 55.75 & 52.91 & 90.33 & 66.73 & 83.88 & 09:03 \\
 & & InstructBLIP & Vicuna-v1.1-7B & 71.63 & \textbf{70.44} & 72.77 & 71.59 & 50.75 & 09:00 \\
 & & Shikra & Vicuna-7B & 73.13 & 67.80 & 86.26 & 75.92 & 62.50 & 28:40 \\
 & & LLaVA-v1.5 & Vicuna-v1.5-7B & 61.50 & 56.10 & \textbf{99.49} & 71.74 & 87.13 & 18:17 \\
 & & GeoChat & Vicuna-v1.5-7B & \textbf{75.63} & 70.04 & 88.04 & \textbf{78.02} & 61.75 & 23:57 \\
 \hline
\multirow{18}{*}{\textbf{Hard}} & \multirow{6}{*}{\textit{Random}} & Minigpt4 & Vicuna-v0-7B & 41.10 & 39.98 & 88.48 & 55.07 & 90.30 & 10:49 \\
 & & Minigpt4 & Llama2-7B & 43.60 & 41.22 & 89.71 & 56.48 & 88.80 & 11:21 \\
 & & InstructBLIP & Vicuna-v1.1-7B & \textbf{78.30} & \textbf{74.30} & 71.57 & \textbf{72.91} & 39.30 & 10:57 \\
 & & Shikra & Vicuna-7B & 68.10 & 57.40 & 84.56 & 68.38 & 60.10 & 35:30 \\
 & & LLaVA-v1.5 & Vicuna-v1.5-7B & 50.30 & 45.17 & \textbf{98.53} & 61.80 & 89.30 & 22:54 \\
 & & GeoChat & Vicuna-v1.5-7B & 71.10 & 59.93 & 87.99 & 71.30 & 59.90 & 31:14 \\
\cline{2-10}
 & \multirow{6}{*}{\textit{Popular}} & Minigpt4 & Vicuna-v0-7B & 44.60 & 41.59 & 88.48 & 56.58 & 86.80 & 10:51 \\
 & & Minigpt4 & Llama2-7B & 50.70 & 44.80 & 89.71 & 59.76 & 81.70 & 11:15 \\
 & & InstructBLIP & Vicuna-v1.1-7B & \textbf{79.80} & \textbf{77.25} & 71.57 & 74.30 & 37.80 & 10:53 \\
 & & Shikra & Vicuna-7B & 76.30 & 66.47 & 84.56 & 74.43 & 51.90 & 35:53 \\
 & & LLaVA-v1.5 & Vicuna-v1.5-7B & 50.30 & 45.02 & \textbf{98.53} & 61.80 & 89.30 & 22:51 \\
 & & GeoChat & Vicuna-v1.5-7B & 77.50 & 67.10 & 88.00 & \textbf{76.14} & 53.50 & 30:47 \\
\cline{2-10}
 & \multirow{6}{*}{\textit{Adversarial}} & Minigpt4 & Vicuna-v0-7B & 44.70 & 41.64 & 88.48 & 56.63 & 86.70 & 10:49 \\
 & & Minigpt4 & Llama2-7B & 50.40 & 44.63 & 89.71 & 59.61 & 82.00 & 11:19 \\
 & & InstructBLIP & Vicuna-v1.1-7B & 77.60 & \textbf{73.00} & 71.57 & 72.28 & 40.00 & 11:03 \\
 & & Shikra & Vicuna-7B & 75.50 & 64.46 & 84.56 & 73.80 & 52.70 & 35:52 \\
 & & LLaVA-v1.5 & Vicuna-v1.5-7B & 50.20 & 44.97 & \textbf{98.53} & 61.75 & 89.40 & 22:54 \\
 & & GeoChat & Vicuna-v1.5-7B & \textbf{79.40} & 69.57 & 87.99 & \textbf{77.71} & 51.60 & 30:48 \\
\hline
\end{tabular}
\vspace{-0.5cm}
\end{table*}



\section{Experiments and Analysis}

\subsection{POPE Metric Evaluation of Remote Sensing Datasets}

Inspired by the POPE evaluation method, we propose a binary classification hallucination evaluation tailored specifically for remote sensing datasets, named RSPOPE. Specifically, we sample the 29 categories of the DDFAV dataset we propose using random, popular, and adversarial sampling methods as defined by POPE. In addition, we innovatively divide the evaluation images into easy, medium, and hard levels to more intuitively and effectively assess the performance of LVLMs. At the easy level, POPE generates 6 binary classification problems per image; at the medium level, it generates 8 binary classification problems per image; and at the hard level, it generates 10 binary classification problems per image. Fig~\ref{fig:RSPOPE} shows part of the RSPOPE evaluation method we designed.

\subsection{Comparison of Zero-Shot Capabilities of Different General LVLMs for Remote Sensing Data}

To demonstrate the versatility of our proposed DDFAV dataset and the RSPOPE evaluation method across different LVLMs, we select 100 images from the DDFAV dataset for each of the easy, medium, and hard levels defined in the RSPOPE evaluation. We then conduct experiments using beam search with the number of beams set to 5. The easy level requires that the picture contains at least 2 different categories of objects, with a total of 5 or fewer objects. The medium level requires at least 3 different categories of objects, with the total number of objects ranging from 6 to 10. The hard level requires at least 4 different categories of objects, with 11 or more objects in the picture. As shown in Table~\ref{general_table}, GeoChat LVLMs achieve the highest F1 score in nine of the RSPOPE evaluation settings, except for the hard-adversarial setting. Additionally, Minigpt4, utilizing Vicuna-v0-7B as the large language model, has the shortest experiment time across all settings, whereas Shikra has the longest.

\section{Conclusion}
This paper analyzed the current situation of remote sensing LVLMs, highlighting the lack of high-quality datasets and reliable hallucination evaluation methods. To address this issue, we proposed a higher-quality remote sensing LVLMs dataset, DDFAV, and produced a training instruction set that could be adapted to various LVLMs tasks. Additionally, we developed a remote sensing LVLMs hallucination evaluation method called "RSPOPE" based on the proposed dataset and conducted zero-shot evaluation experiments on multiple LVLMs. We plan to further expand the scale of the training instruction set and evaluation method in the future.

{\tiny
\bibliography{ref}
}
\vspace{12pt}

\end{document}